\documentclass[10pt, conference]{IEEEtran}
\IEEEoverridecommandlockouts
\AtBeginDocument{
  }
\usepackage{url}
\usepackage{bm}
\usepackage{amsmath}
\usepackage{amssymb}
\usepackage{mathtools}
\usepackage{soul}
\usepackage{subcaption}
\usepackage{algorithm}
\usepackage{algpseudocode}
\usepackage[utf8]{inputenc}
\usepackage{graphicx}
\usepackage{cuted}
\usepackage{capt-of}
\usepackage{comment}
\usepackage{caption}
\usepackage[dvipsnames]{xcolor}

\begin{document}

\title{SafeStep: An Interactive Demonstration of Semantic Communication for Pedestrian Safety Monitoring\\
\thanks{This work was supported in part by the NSF under Grant CNS-2239677, the Alabama Research and Development Enhancement Fund (ARDEF) program under Grant Agreement No. 1ARDEF26 11,  a donation from NVIDIA, and Auburn University.}
}
\author{
\IEEEauthorblockN{Christian McDowell\textsuperscript{\(\dagger\)}, Andrea Panebianco\textsuperscript{\(\dagger\)}, Jeremiah Yang\textsuperscript{\(\dagger\)}, \\Sirin Chakraborty, Samuel Chamoun, Travis Ross, Yin Sun\\
{\textit{Department of Electrical and Computer Engineering, Auburn University, Alabama, USA}}\\
%{\textit{Email:} \{cdm0135, anp0105, jjy0006, szc0260, sgc0042, trp0021, yzs0078\}@auburn.edu}\\
{\textsuperscript{\(\dagger\)}\textit{Co-primary authors.}}}
}

\maketitle

\begin{abstract}
%NextG edge-intelligence systems must preserve task-relevant information as channel quality, symbol budget, and information age change. Offline transceiver scores, however, do not show how these conditions alter the receiver view of an evolving scene during live multiuser operation. 
In this paper, we develop SafeStep, an interactive browser-based semantic communication platform for live pedestrian safety monitoring. SafeStep extracts pedestrian information from four live traffic-camera feeds, transmits it through a semantic communication
%user-selected
transceiver over a software-emulated Additive White Gaussian Noise (AWGN) channel, and renders user-specific positions, trajectories, and risk labels. The platform 
%processes each shared scene update once and 
allows each user to select the transceiver, Signal-to-Noise Ratio (SNR), codelength, and Age of Information (AoI) and view the resulting pedestrian reconstruction. SafeStep compares a recently proposed semantic communication design called Meta-VIB with five baseline transceivers.
Meta-VIB uses a compact neural model with only $4.16$ million parameters to generalize across varying SNR, codelength, and AoI values without online retraining.
%Meta-VIB uses one model with SNR- and codelength-dependent Feature-wise Linear Modulation, ordered-prefix transmission, and AoI-aware decoding to operate across the evaluated conditions without online retraining.
Meta-VIB achieves mean task-loss reductions of up to $92.1\%$. On one high-end GPU server, the integrated concurrent-access workload maintains the target $5~\mathrm{frames/s}$ through $20$ users. At $100$ users, each requesting a distinct configuration, SafeStep records no request failures and a mean application response time below $1~\mathrm{s}$, but its mean per-browser frame rate falls to approximately $1~\mathrm{frame/s}$. 
%SafeStep therefore moves semantic communication evaluation beyond offline aggregate metrics by making both transceiver performance and application responsiveness directly observable during live multiuser operation. 
To our knowledge, SafeStep is the first real-time semantic communication platform to make AoI-induced downstream degradation directly observable in live monitoring applications.
\end{abstract}

\begin{comment}
\begin{abstract}
In this paper, we develop SafeStep, an interactive browser-based semantic communication platform for live pedestrian safety monitoring. SafeStep extracts pedestrian information from four live traffic-camera feeds, transmits it through a semantic communication transceiver over an additive white Gaussian noise (AWGN) channel, and renders user-specific positions, trajectories, and risk labels. Through a browser, each user independently selects the transceiver, signal-to-noise ratio (SNR), codelength, and age of information (AoI) and observes the resulting pedestrian reconstruction. SafeStep compares Meta-VIB, a recently proposed semantic communication design, with five baseline transceivers. Meta-VIB uses a compact neural model with only $4.16$ million parameters to generalize across varying SNR, codelength, and AoI values without online retraining. Experimental results show that Meta-VIB achieves mean task-loss reductions of up to $92.1\%$. On one high-end GPU server, the integrated concurrent-access workload maintains the target $5~\mathrm{frames/s}$ through $20$ users. At $100$ users, each requesting a distinct configuration, SafeStep records no request failures and a mean application response time below $1~\mathrm{s}$, but its mean per-browser frame rate falls to approximately $1~\mathrm{frame/s}$. To our knowledge, SafeStep is the first real-time semantic communication platform to make AoI-induced downstream degradation directly observable in live monitoring applications.
\end{abstract}

\end{comment}
\section{Introduction}

NextG edge-intelligence systems integrate edge sensors, neural-network-based
%learned 
wireless transceivers, and downstream inference or control loops. In safety-critical applications, the communication objective is not to reconstruct every detail of the raw sensor data, but to preserve the semantic information most relevant to downstream safety decisions under channel noise and limited codelength.
%, and information freshness constraints.
Task-oriented communication preserves information needed for a downstream task~\cite{gunduz2022beyond,kountouris2021semantics,qiao2025todma}. Information Bottleneck (IB) methods balance compression with task relevance. Variational Information Bottleneck (VIB) methods learn compact stochastic representations that retain task-relevant information~\cite{alemi2016deep,peng2025hyper}.
%, and freshness-aware communication based on Age of Information (AoI)~\cite{sun2017update,yates2021age}
However, existing systems are typically evaluated
%offline 
through
%aggregate
averaging performance metrics, such as throughput, delay, and inference accuracy~\cite{bourtsoulatze2019deep,diao2025aligning,xu2023deep}. Such evaluations do not directly reveal how channel noise and codelength
%, and outdated information
affect the receiver-side reconstruction of the 
%same 
evolving safety-critical scene during live operation.

Recently, we introduced Meta-VIB~\cite{mcdowell2026significance}, a semantic communication design with several technical innovations: (i) it is the first Joint Source--Channel Coding (JSCC)-based semantic communication design to support Age of Information (AoI), establishing a connection between AoI and physical-layer semantic communication that has not been explicitly established in prior work; (ii) it introduces a per-sample data-importance metric, together with a method for evaluating it, grounded in both Bayesian decision theory and information theory; and (iii) one model generalizes across Signal-to-Noise Ratio (SNR), AoI, and codelength values without online retraining. Feature-wise Linear Modulation (FiLM)~\cite{perez2018film} conditions the network on SNR and codelength, while AoI separately conditions the decoder. This conditioning can support generalization to values not used during training. In addition, Meta-VIB incorporates an information-concentration regularizer to 
%enable 
realize dynamic codelength truncation.

In this paper, we further develop \textbf{SafeStep}\footnote{\url{https://safestep.eng.auburn.edu/}}, a web-based, live, and interactive demonstration of Meta-VIB and several other semantic communication designs in the context of pedestrian safety monitoring. SafeStep uses four live traffic-camera feeds from Toomer's Corner, a 
%well-known
busy intersection in Auburn, Alabama, to demonstrate how safety-critical pedestrian information can be transmitted and reconstructed over a software-emulated
Additive White Gaussian Noise (AWGN) channel. The web interface provides synchronized transmitter and receiver views, allowing users to observe the original traffic scene alongside the reconstructed pedestrian positions and estimated safety statuses. Through a browser, users can interactively select different semantic communication
%designs
transceivers and communication conditions to examine how SNR, codelength, and AoI affect the preservation of safety-critical information.

\begin{figure*}[t]
\centering
\includegraphics[width=0.811\textwidth]{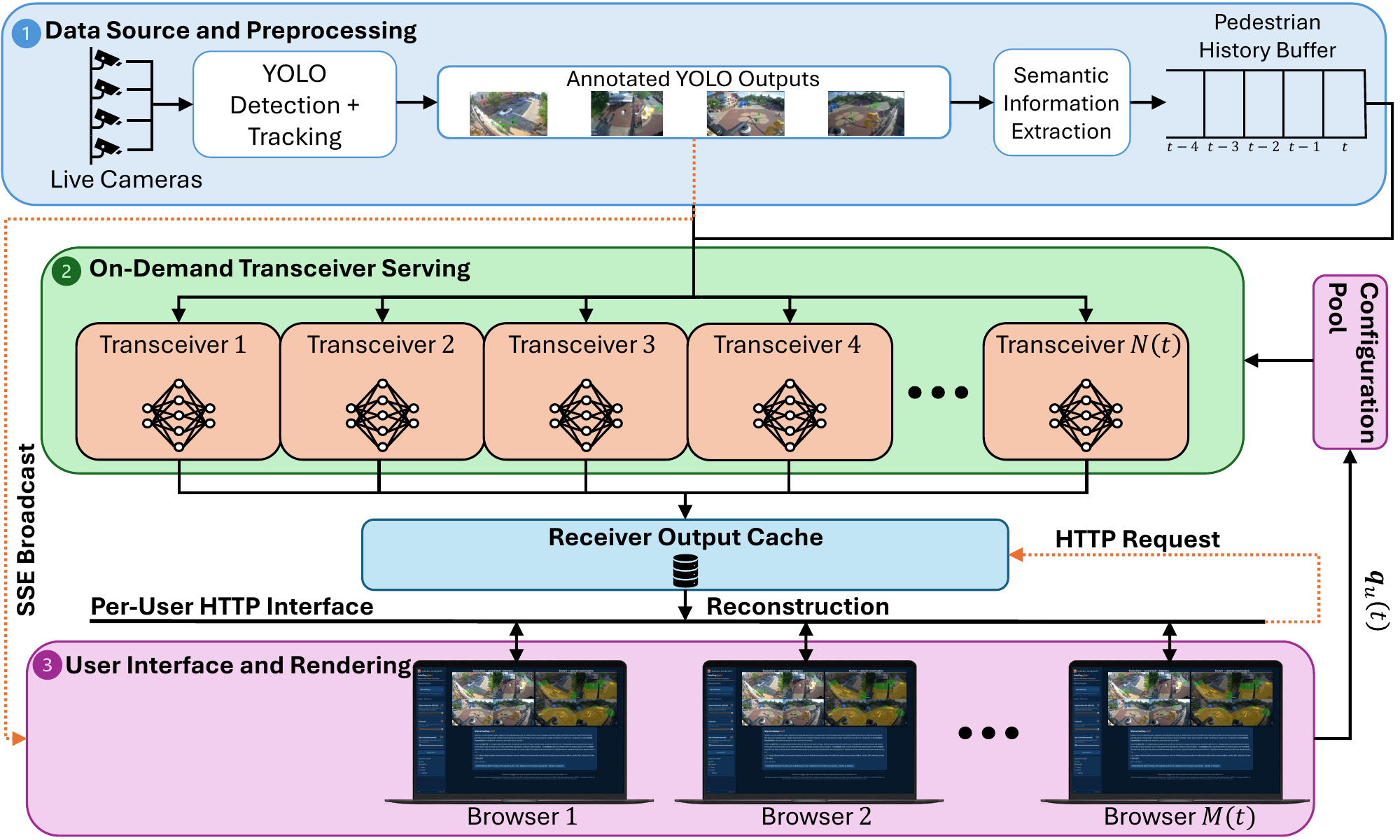}
\caption{SafeStep system design block diagram. 
%Shared live processing is performed once per frame and reused across users, while each browser independently selects its semantic communication design and communication conditions and receives a user-specific receiver-side reconstruction.
}
\label{fig:system}
\end{figure*}

The technical contributions of this paper are summarized as follows:
\begin{itemize}
\item \textbf{Interactive visualization and concurrent access:}
SafeStep provides an interactive visualization of reconstructed pedestrian positions and the corresponding safety statuses (\emph{safe}, \emph{cautious}, or \emph{dangerous}). It allows users to compare different semantic communication
%designs 
transceivers over a wide range of communication conditions, including SNR values of $-10,-9,\ldots,20~\mathrm{dB}$, codelengths of $2,3,\ldots,16$ complex Gaussian channel symbols, and AoI values ranging from $0$ to $6$ seconds. SafeStep supports multiple concurrent users, each of whom can independently select a semantic communication 
%design 
transceiver and communication conditions. A single GPU server can simultaneously handle inference requests from at least $100$ different configurations of semantic communication
%design
transceiver, SNR, codelength, and AoI. The architecture enabling this large-scale concurrent access is presented in Section~\ref{sec:design}.
\item \textbf{Generalization and system performance:}
SafeStep evaluates transceiver generalization and application-level responsiveness. Meta-VIB reduces mean task loss by up to $92.1\%$ relative to the worst-performing baseline at the same setting (Fig.~\ref{fig:generalization}). In the integrated concurrent-access experiment, each browser requests a distinct configuration. On one high-end GPU server, SafeStep maintains the target rate of $5~\mathrm{frames/s}$ through $20$ concurrent users. At $100$ users, it records no request failures and keeps the mean reconstruction response time below $1~\mathrm{s}$, but its mean per-browser frame rate falls to approximately $1~\mathrm{frame/s}$, as shown in Fig.~\ref{fig:stress}. 
%During the concurrent-access experiment, SafeStep observed no request failures, and the maximum application response delay remained below $1$ second, as seen in Fig.~\ref{fig:stress}. SafeStep maintained the target $5~\mathrm{frame/s}$ through $20$ users. However, the mean per-browser frame rate falls to approximately $1~\mathrm{frame/s}$.

\end{itemize}

\section{Related Work}
\label{sec:related}

SafeStep draws on learned Joint Source--Channel Coding (JSCC) and task-oriented representation learning. DeepJSCC jointly trains an encoder and decoder around a noisy channel~\cite{bourtsoulatze2019deep}. Adaptive variants support changing SNR and bandwidth, while DeepJSCC-$\ell$++ supplies both quantities to its encoder and decoder during training and inference~\cite{xu2023deep,bian2023deepjscclpp}. VIB, Hyper-VIB, and ATROC learn compressed representations for downstream tasks~\cite{alemi2016deep,peng2025hyper,diao2025aligning}. These methods do not use AoI as a neural-network input.

Prior work relates freshness to learned communication without conditioning the neural networks on age. Sagduyu \emph{et al.} define peak Age of Task Information, which resets only after a correct classification, and use it to adjust the number of channel uses~\cite{sagduyu2023age}. Basnayaka \emph{et al.} define Age of Misclassified Information and use its average to evaluate a JSCC image-classification system~\cite{basnayaka2024freshness}. In both studies, age is computed outside the encoder and receiver model; it is neither a neural-network input nor a training loss. Meta-VIB instead trains over multiple age offsets~\cite{mcdowell2026significance}. Its encoder does not receive AoI, while its decoder receives the discrete age offset during training and inference. The decoder can therefore adjust its prediction to the age of the available pedestrian history.

Hardware prototypes have demonstrated image reconstruction on a field-programmable gate array over a software-defined radio link and modular over-the-air testbeds that combine software-defined radios with edge graphics processors~\cite{yoo2022demo,ding2026prototyping}. SafeStep studies a complementary application-level setting in which concurrent browser users observe how SNR, codelength, and AoI affect live pedestrian reconstructions over a software-emulated AWGN channel.

\begin{figure}[t]
\centering
\includegraphics[width=\columnwidth]{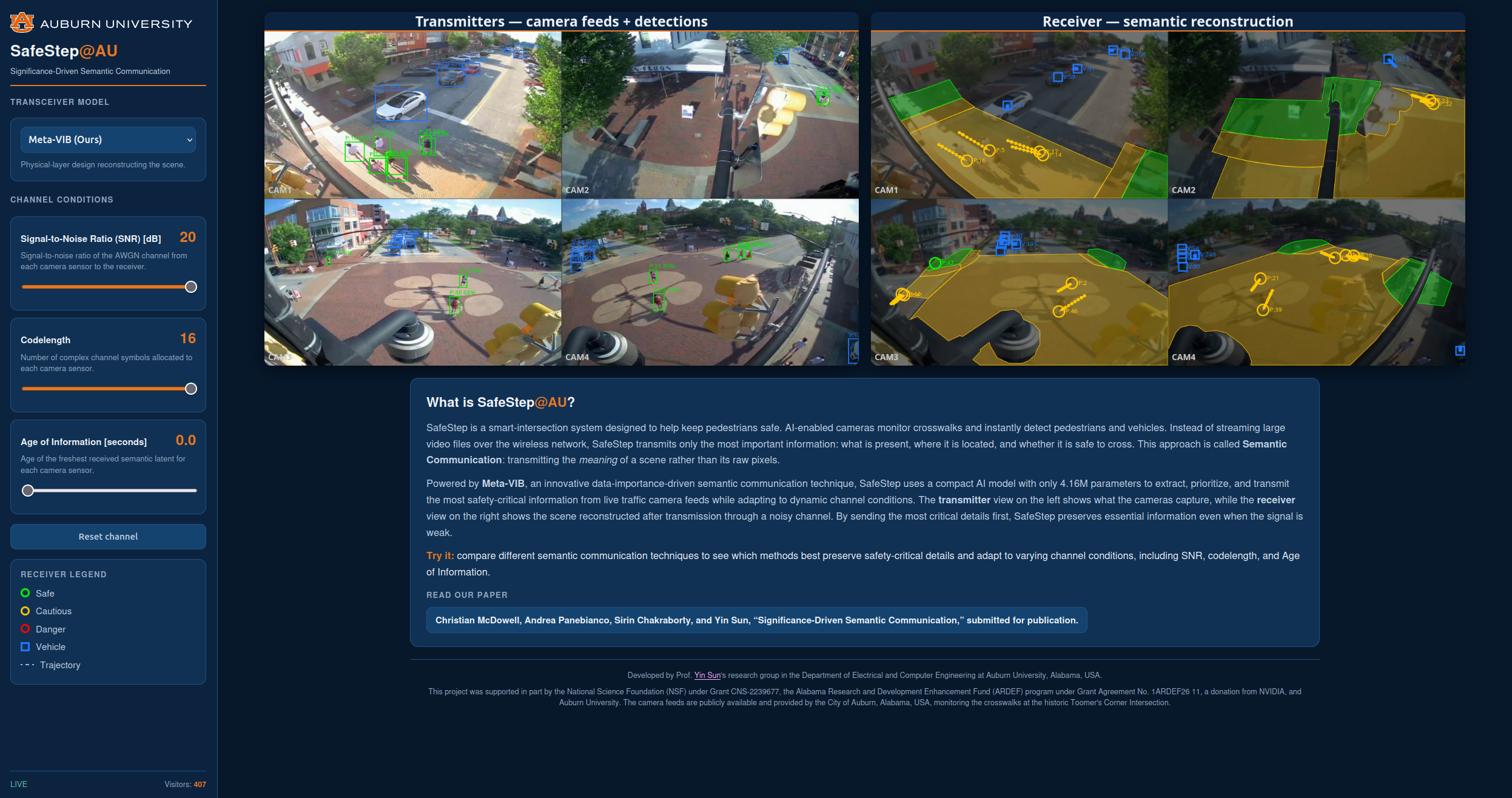}
\caption{SafeStep browser interface. 
%The transmitter view displays the shared live scene, while the receiver view displays the pedestrian reconstruction produced under the user's selected transceiver, SNR, codelength, and AoI. %The risk zones come from the shared preprocessing pipeline and are not reconstructed by the transceiver.
}
\label{fig:ui}
\end{figure}

\section{SafeStep System Design}
\label{sec:design}

SafeStep consists of three parts, as shown in Fig.~\ref{fig:system}: (i) data source and preprocessing, (ii) on-demand transceiver serving, and (iii) user interface and rendering. These parts allow many users to explore the same live scene without repeating the entire processing pipeline for every browser.

The first part processes the four live camera views once. Each newly processed set of camera frames is called a \emph{source update}. A source update contains the shared scene information and the recent history of each tracked pedestrian. The second part uses these pedestrian histories to run the requested transceivers. When several requests require the same computation, the server computes the reconstruction once and reuses it. The third part combines the shared scene information with the requested pedestrian reconstruction inside each browser. The platform therefore performs shared processing once per source update, transceiver processing once per required computation, and rendering once per browser.

%SafeStep must serve concurrent browsers that observe the same evolving traffic scene but may request different transceiver designs and communication conditions. Its central design objective is therefore to share source-side work while preserving configuration-specific communication effects. As shown in Fig.~\ref{fig:system}, the platform has three components: (i) data source and preprocessing, (ii) on-demand transceiver serving, and (iii) user interface and rendering. These components perform work once per source update, once per distinct effective transceiver computation, and once per browser, respectively.

%To distinguish delivery fan-out from computational diversity, let $\mathbf{q}_u(t)=(m_u(t),\zeta_u(t),\eta_u(t),\Delta_u(t))$ denote user $u$'s selected transceiver design, SNR, requested codelength, and AoI at time $t$. The platform serves $M(t)$ connected users and $N(t)=|\{\mathbf{q}_u(t):u=1,\ldots,M(t)\}|\leq M(t)$ distinct requested configurations. The number of effective transceiver computations can be smaller than $N(t)$ when different configurations invoke the same operation.

\subsection{Data Source and Preprocessing}
\label{sec:design:data}

SafeStep begins with four live views of Toomer's Corner, a busy signalized intersection in Auburn, Alabama. The four views arrive through one public YouTube stream as a $2{\times}2$ traffic-camera mosaic~\cite{toomers2024webcam}. SafeStep uses \texttt{yt-dlp} to locate the current stream and \texttt{ffmpeg} to decode it continuously. The platform separates each decoded frame into four camera views and processes those views once per source update.

SafeStep detects vehicles and pedestrians before running any transceiver. We created a custom labeled dataset and trained separate YOLO26l vehicle and YOLO11s pedestrian detectors~\cite{ultralytics2026yolo26}. The vehicle detections help determine whether traffic is stopped or moving through the intersection; they are not transceiver inputs. The pedestrian detections provide the position observations used by the transceivers. A greedy tracker links pedestrian detections across frames using intersection over union (IoU), which measures the overlap between two bounding boxes. The tracker assigns the same identifier to detections that belong to the same pedestrian and uses them to build a short motion history. This simple overlap-based tracker keeps the live pipeline lightweight for fixed cameras and short histories.

Each pedestrian position is represented by the center of its bounding box, called the \emph{centroid}. The detector-derived centroid serves as the reference position used by SafeStep. The live camera frames are not manually labeled, so these reference positions may contain missed detections or localization errors.
%They should not be interpreted as ground-truth pedestrian positions.

%SafeStep next converts the detected scene into information for the pedestrian-safety task. A \emph{risk zone} is a fixed polygonal region assigned one of three labels: \emph{safe}, \emph{cautious}, or \emph{danger}. SafeStep estimates the traffic-signal state from predefined image regions and uses observed vehicle motion to update these labels. Sidewalks and waiting areas remain \emph{safe}. When traffic is stopped, crosswalks and curb-adjacent areas are labeled \emph{cautious}. When vehicles are moving, roadways and crosswalks are labeled \emph{danger}. Each pedestrian receives the label of the risk zone containing its centroid. These labels estimate traffic risk; they do not represent verified safety outcomes.

SafeStep next converts the detected scene into information for the pedestrian-safety task. A \emph{risk zone} is a fixed polygonal region assigned one of three labels: \emph{safe}, \emph{cautious}, or \emph{dangerous}. SafeStep estimates the traffic-signal state from predefined image regions and uses observed vehicle motion to update these labels. These zone definitions are further explained in Appendix D of \cite{mcdowell2026significance}. Each pedestrian receives the label of the risk zone containing its centroid.

SafeStep normalizes each centroid by the width and height of its camera view. Every $T_s=0.2~\mathrm{s}$, where $T_s$ is the sampling interval, the platform records each pedestrian's normalized centroid and risk label. The five most recent records form the pedestrian's \emph{semantic history}. The encoder converts this history into complex channel symbols, and only these symbols pass through the emulated AWGN channel. The decoder uses the received symbols to reconstruct the pedestrian's positions, trajectory, and risk-label probabilities. Camera images, vehicle and pedestrian detection boxes, and risk zones do not pass through the emulated channel; SafeStep sends them separately as shared rendering information.

%SafeStep records the tracked scene every $T_s=0.2~\mathrm{s}$, where $T_s$ is the sampling interval. Each pedestrian record contains a track identifier, a normalized centroid, and a risk label. A normalized centroid expresses the bounding-box center relative to the width and height of its camera view. Up to five recent records for one pedestrian form its \emph{semantic history}. The transceivers process these compact histories rather than the camera images.

%The camera images, detection boxes, and risk zones form the shared rendering context sent separately to the browsers. This context does not pass through the emulated wireless channel. SafeStep produces the semantic histories and rendering context once per source update. Additional users and AoI selections therefore do not repeat stream decoding, object detection, or tracking.

\begin{figure*}[t]
\centering
\includegraphics[width=.879\textwidth]{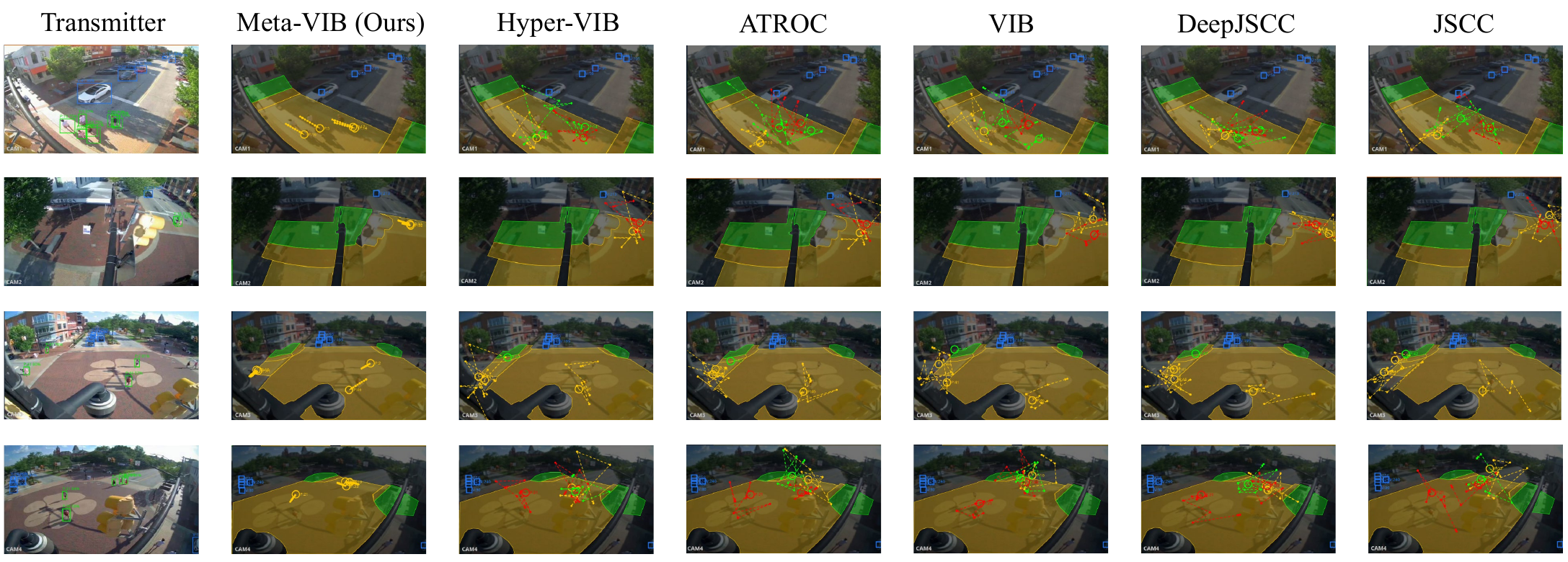}
\caption{Shared transmitter and receiver views from all transceivers at $\zeta=0~\mathrm{dB}$, codelength $\eta=8$, and $\Delta=0.8~\mathrm{s}$. 
%The figure shows how one live scene appears under each deployed design. 
%Because the designs use different transmitted widths and AoI mechanisms, this is a qualitative platform demonstration rather than an equal-symbol-budget comparison.
}
\label{fig:mosaic}
\end{figure*}

\subsection{On-Demand Transceiver Serving}
\label{sec:design:txrx}

The live scene is shared, but each browser can request a different transceiver, SNR, codelength, and AoI. 
%and different communication conditions. 
Running a separate model pipeline for every browser would repeat work when users make the same request. SafeStep avoids this duplication by sharing pretrained models and reusing reconstructions.

At time $t$, user $u$ requests the configuration
\begin{align}
\mathbf{q}_u(t)
&=
\bigl(m_u(t),\zeta_u(t),\eta_u(t),\Delta_u(t)\bigr),
\label{eq:user_configuration}
\end{align}
where $m_u(t)$ identifies the transceiver. The variables $\zeta_u(t)$, $\eta_u(t)$, and $\Delta_u(t)$ specify the SNR, requested codelength, and AoI, respectively. Let $M(t)$ denote the number of connected browsers. Let $N(t)\leq M(t)$ denote the number of different configurations requested by those browsers.

%A requested configuration describes how SafeStep uses a model; it does not create a new copy of that model.
SafeStep keeps one pretrained model for each of the six supported designs loaded in GPU memory. The memory used for model parameters therefore remains fixed as $M(t)$ increases.

Different configurations do not always require different server computations. SafeStep calls the computation that remains after ignoring settings unsupported by the selected model an \emph{effective operation}. Meta-VIB uses the selected SNR, codelength, and AoI during model computation. However, baselines use the selected SNR to set the channel noise, but their requested codelength and AoI do not change their server-side model computation. The number of distinct effective operations can therefore be smaller than $N(t)$. Thus, $N(t)$ counts requested configurations, not model copies or transceiver executions.
%Several configurations can therefore require the same effective operation. Thus, $N(t)$ counts requested configurations, not model copies or neural-network executions.

SafeStep also reuses completed results. Before running a transceiver, the server checks a cache indexed by the source update and effective operation. If the requested result is already stored, the server returns it immediately. Otherwise, the server runs the transceiver, applies the AWGN channel, decodes the received representation, and stores the reconstruction until the next source update.

%The cache stores the result after channel noise and decoding have been applied. Matching requests during the same source update therefore receive the same reconstruction and the same AWGN noise realization. These requests do not represent independent simulated wireless links. This design avoids repeating the same model computation for identical requests.

\subsubsection{Supported Semantic Communication Designs}
\label{sec:design:rx}

The serving layer supports six transceiver designs through the same input and output format. SafeStep provides Meta-VIB~\cite{mcdowell2026significance} and five baselines: JSCC~\cite{bourtsoulatze2019deep}, DeepJSCC~\cite{xu2023deep}, VIB~\cite{alemi2016deep}, Hyper-VIB~\cite{peng2025hyper}, and ATROC~\cite{diao2025aligning}. Each design receives the same pedestrian history. Each design returns five estimated pedestrian centroids and a risk-label distribution for each estimate. SafeStep can therefore change the transceiver without changing the preprocessing, serving, or rendering pipelines.

The five baselines represent different ways to learn a compact transmitted representation. JSCC and DeepJSCC learn direct mappings from source information to channel symbols and receiver outputs. VIB learns a stochastic representation that preserves information needed for the task. Hyper-VIB uses a hypernetwork to adjust the VIB model. ATROC preserves information for both source reconstruction and the downstream task. SafeStep applies each approach to the same pedestrian-safety task.

Meta-VIB differs from the baselines because its model
%computation changes 
generalizes across
the selected communication conditions. 
Meta-VIB first uses a recurrent network to summarize each pedestrian history. SafeStep computes this summary once for each history and reuses it across Meta-VIB requests during the same source update. A lightweight hypernetwork then generates the scale and shift values used by FiLM~\cite{perez2018film}. FiLM applies these values to intermediate network features, enabling the encoder and decoder to adapt to the selected SNR and codelength. 
%The selected SNR and codelength condition both the encoder and decoder. The selected AoI also conditions the decoder.

%The Meta-VIB encoder produces a compact representation with a maximum length of $K=16$ coordinates. 
The Meta-VIB encoder produces an ordered representation of $K=16$ complex channel symbols. Training encourages the most useful task information to appear near the beginning of this representation. For requested codelength $\eta_u(t)\leq K$, Meta-VIB retains the first $\eta_u(t)$ coordinates. The retained coordinates are normalized to meet the transmission-power constraint, then transmitted over the AWGN channel at the selected SNR. The receiver places the noisy coordinates in the first $\eta_u(t)$ decoder inputs and fills the remaining $K-\eta_u(t)$ inputs with zeros.

SafeStep converts the selected AoI into the discrete age offset $\delta_u(t)=\Delta_u(t)/T_s$, which counts the number of sampling intervals in the selected AoI. Meta-VIB is trained using multiple age offsets, and its decoder receives the selected $\delta_u(t)$ at inference. Relative to the newest observation in the pedestrian history, the five prediction targets lie $\delta_u(t)+1,\ldots,\delta_u(t)+5$ samples ahead. The age offset therefore tells the decoder how far into the future it must predict.

The baseline networks use the same task interface, but they do not receive SNR, requested codelength, or AoI as inputs. For baseline model $m$, $K_m$ denotes the number of complex channel symbols that the model was trained to produce. Thus, $K_m$ is a training choice, whereas $\eta$ is the codelength requested at run time. The live platform uses a model with $K_m=2$ for each baseline. Because the interface restricts $\eta_u(t)\geq2$, every live baseline transmits both symbols. The selected AoI affects only the live browser rendering described in Section~\ref{sec:design:ui}.

Section~\ref{sec:eval:generalization} also evaluates separately trained baseline models with $K_m=8$. When a baseline is tested at requested codelength $\eta$, it transmits the first $\min\{\eta,K_m\}$ symbols produced by its encoder. This test-time truncation can shorten the transmitted representation, but the encoder does not adapt its representation to $\eta$. The controlled AoI evaluation also advances the prediction target as information ages.
%, but the baseline decoders still do not receive the age offset.

%The baseline models use the same input and output format as Meta-VIB, but their networks do not receive SNR, requested codelength, or AoI as inputs. For baseline $m$, let $K_m$ denote the training codelength of its selected model. The live platform loads a model trained to produce $K_m=2$ complex channel symbols for each baseline. Because the selectable codelength satisfies $\eta_u(t)\geq2$, each model always transmits two complex symbols. The selected SNR changes only the AWGN variance. The selected AoI does not enter the baseline network and changes only the live browser rendering described in Section~\ref{sec:design:ui}. Section~\ref{sec:eval:generalization} evaluates the deployed $K_m=2$ models and additional models trained to produce $K_m=8$ symbols. At requested codelength $\eta$, a baseline transmits the first $\min\{\eta, K_m\}$ symbols produced by its encoder. The requested codelength can therefore shorten the transmitted prefix, but the baseline network does not adapt its encoding to $\eta$.

\begin{figure*}[t]
\centering
\begin{subfigure}{0.32\textwidth}
\centering
\includegraphics[width=\linewidth]{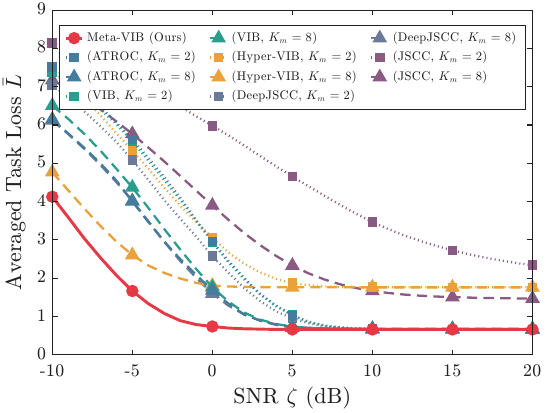}
\caption{$\bar L$ versus SNR at $\eta=8$ and $\Delta=0.8~\mathrm{s}$.}
\label{fig:generalization:snr}
\end{subfigure}
\hfill
\begin{subfigure}{0.32\textwidth}
\centering
\includegraphics[width=\linewidth]{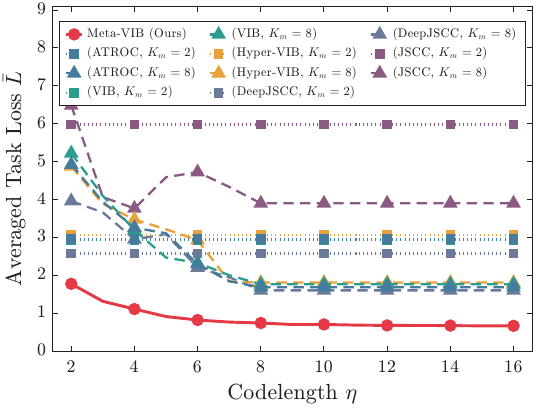}
\caption{$\bar L$ versus $\eta$ at $\zeta=0~\mathrm{dB}$ and $\Delta=0.8~\mathrm{s}$.}
\label{fig:generalization:codelength}
\end{subfigure}
\hfill
\begin{subfigure}{0.325\textwidth}
\centering
\includegraphics[width=\linewidth]{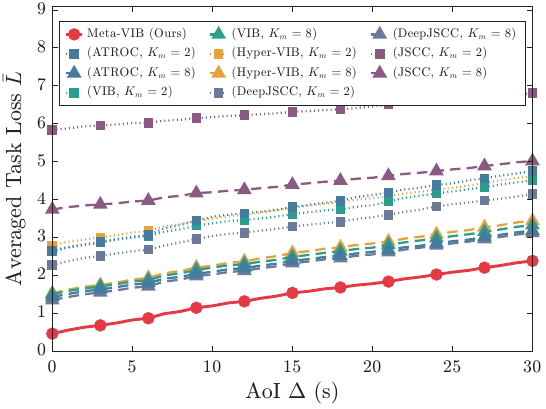}
\caption{$\bar L$ versus AoI at $\zeta=0~\mathrm{dB}$ and $\eta=8$.}
\label{fig:generalization:aoi}
\end{subfigure}
\caption{Mean task loss $\bar L$ under varying SNR, codelength, and AoI. Baselines use separate $K_m\in\{2,8\}$ models.
%Color and marker identify the transceiver, while line style identifies the trained baseline width $K_m\in\{2,8,16\}$. 
%Panels~(a) and~(c) use $K_m=8$ for the primary equal-budget comparison with Meta-VIB at $\eta=8$. Panel~(b) uses the matched points $\eta=K_m$ for the primary codelength comparison. Lower $\bar L$ is better.
}
\label{fig:generalization}
\end{figure*}

\subsection{User Interface and Rendering}
\label{sec:design:ui}

The browser turns the shared live data and the requested transceiver output into an interactive view. When a browser opens, it begins with a default configuration of $\mathbf{q}_u(t)=(\text{Meta-VIB}, 20~\mathrm{dB}, 16, 0~\mathrm{s})$ for user $u$. The user can then change the transceiver, SNR, codelength, or AoI without changing another user's view. As shown in Fig.~\ref{fig:ui}, the transmitter view displays the shared camera feeds and perception results. The receiver view displays the pedestrian positions, trajectories, and risk labels produced under the user's selected settings. The risk zones appear in both views, but they come from the shared preprocessing pipeline described in Section \ref{sec:design:data} and do not pass through the emulated wireless channel.

SafeStep uses two delivery paths to build these views. 
%Server-Sent Events (SSE) send each shared scene update to every connected browser. 
Server-Sent Events (SSE) broadcast each shared scene update to all connected browsers.
After a source update or control change, browser $M$ sends its users requested configuration $\mathbf{q}_u(t)$ to the server
%\texttt{/semantic} endpoint 
through a Hypertext Transfer Protocol (HTTP) request. The server then returns the decoded centroid sequence and risk-label probabilities.
%in JavaScript Object Notation (JSON) format. 
The browser then draws the reconstructed positions, trajectory, and risk label over the shared scene. This design sends the live scene once and avoids generating a separate video stream for every user.

%Three bounded buffers keep the interface close to the live scene while supporting the AoI control. The server stores at most $10$ scene updates. Each SSE connection has a queue that holds at most $4$ updates waiting for delivery. If a browser falls behind, the queue discards older updates and keeps the newest scene information. This rule prevents network delay from creating an increasingly old display.

Each browser also stores its $75$ most recent scene updates. At the nominal rate of $5$ updates per second, this history covers about $15~\mathrm{s}$. The history is therefore long enough to support the complete selectable AoI range of $0$ to $6~\mathrm{s}$. The server buffer and SSE queue 
keep delivery current, while the longer browser history supports age-based rendering.

Before drawing a reconstructed trajectory, SafeStep aligns it with the live camera view. The browser places the first decoded point at the pedestrian's current 
%detector-derived 
position. It then preserves the relative movement between the remaining decoded points. 
SafeStep applies this alignment to all six transceivers. 
The detector-derived position used for alignment is side information and does not pass through the emulated channel.

%The browser displays AoI differently for Meta-VIB and the baselines. Let $J$ denote the index of the newest scene update stored by the browser. For a baseline, the browser displays the scene update $i_u(t)=\max\{0, J-\delta_u(t)\}$.
%This update shows the scene that was available $\delta_u(t)$ sampling intervals earlier. If the requested update is not available during startup, the browser uses the oldest update currently stored. This selection changes only the displayed background.
%; the baseline transceiver does not receive the age offset as an input.
%Meta-VIB instead keeps the current scene as its background. 
%The server supplies $\delta_u(t)$ to the Meta-VIB decoder so that the decoder can account for the age of the available pedestrian history.
%To make the selected age visible on the current background, the browser also moves the displayed pedestrian marker backward along its observed motion. 
%This back-projection occurs only in the visualization. It does not change the decoded trajectory or the detector-derived position used to align that trajectory.

The browser applies the same AoI-rendering procedure to all six transceivers. Let $J$ denote the index of the newest scene update stored by the browser. For browser session $u$, the browser displays update $i_u(t)=\max\{0,J-\delta_u(t)\}$, which shows the scene available $\delta_u(t)$ sampling intervals earlier. If this update is unavailable during startup, the browser displays the oldest stored update. The browser also moves the displayed pedestrian marker backward along its observed motion to make the selected age visible. These browser-side changes affect only the visualization. Meta-VIB separately receives $\delta_u(t)$ as a decoder input, whereas the baseline decoders do not.

%These rendering rules show how each deployed design handles AoI in the live interface. They are separate from the controlled AoI experiment in Section~\ref{sec:eval:generalization}. The controlled experiment does not use an older background or move a displayed marker. Instead, it evaluates every transceiver against the same target as the available pedestrian history becomes older.

\begin{figure*}[t]
\centering
\begin{subfigure}{0.32\textwidth}
\centering
\includegraphics[width=\linewidth]{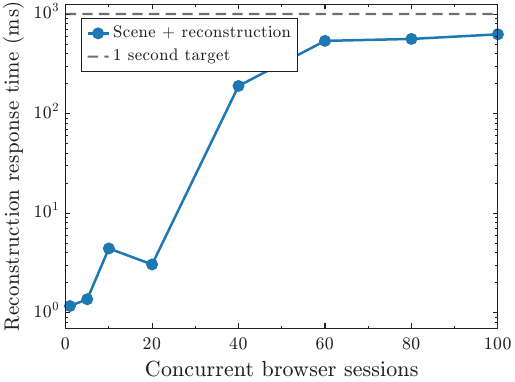}
\caption{Reconstruction response time.}
\label{fig:stress:delay}
\end{subfigure}
\hfill
\begin{subfigure}{0.327\textwidth}
\centering
\includegraphics[width=\linewidth]{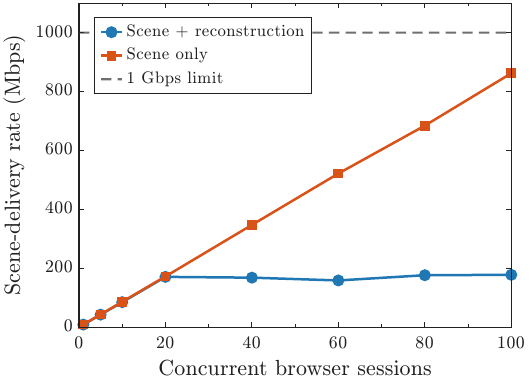}
\caption{Scene-delivery rate.}
\label{fig:stress:throughput}
\end{subfigure}
\hfill
\begin{subfigure}{0.31\textwidth}
\centering
\includegraphics[width=\linewidth]{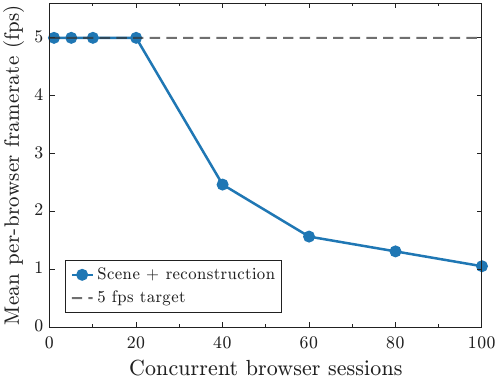}
\caption{Mean per-browser framerate.}
\label{fig:stress:fps}
\end{subfigure}
\caption{Stress test performance versus concurrent browser sessions.
%Concurrent-access performance on a server equipped with an NVIDIA RTX PRO 6000 Blackwell Max-Q Workstation Edition GPU. The plots compare SSE-only delivery with the integrated SSE+HTTP workload as the number of users increases. 
%Each integrated user selects a distinct requested configuration, and the target scene-delivery rate is $5$ frames per second.
}
\label{fig:stress}
\end{figure*}

\section{Evaluation}
\label{sec:eval}

The evaluation follows SafeStep from the user interface to the transceiver performance and then to the server capacity. Fig.~\ref{fig:mosaic} shows how the selected transceiver changes the pedestrian positions, trajectories, and risk labels displayed in the receiver view. Fig.~\ref{fig:generalization} compares the transceivers as SNR, codelength, and AoI change. Fig.~\ref{fig:stress} shows how the complete platform responds as the number of users increases.

\subsection{Performance Across Communication Conditions}
\label{sec:eval:generalization}

Fig.~\ref{fig:mosaic} shows a qualitative shared transmitter view and the six receiver views captured at the same interface instant. 
%The selected settings are $\zeta=0~\mathrm{dB}$, requested codelength $\eta=8$, and AoI $\Delta=0.8~\mathrm{s}$. 
The differences among the reconstructed pedestrian positions, trajectories, and risk labels make the effect of the selected transceiver visible to the user.

Fig.~\ref{fig:generalization} plots the mean task loss $\bar L$ versus SNR $\zeta$, requested codelength $\eta$, and AoI $\Delta$. Each point averages the task loss defined in~\cite{mcdowell2026significance} over cameras, pedestrians, and the five-step prediction horizon; lower $\bar L$ indicates better preservation of the positions, trajectories, and risk labels needed by the safety task. The panels include separate baseline models trained at codelengths $K_m\in\{2,8\}$, while Meta-VIB uses one model throughout. In Figs.~\ref{fig:generalization:snr} and~\ref{fig:generalization:aoi}, Meta-VIB and the $K_m=8$ baselines transmit eight symbols; the $K_m=2$ baselines transmit two.
%, and baseline $m$ transmits $\min\{\eta,K_m\}$ symbols.
Fig.~\ref{fig:generalization:codelength} instead sweeps $\eta$.
%, so the remaining baseline points show truncation or fixed-width behavior. 
One can observe that $\bar L$ generally decreases as SNR or codelength increases and rises as AoI increases. Meta-VIB produces the lowest plotted mean over most of the SNR range and provides its clearest gains when channel noise is substantial. It also performs well over several codelengths using one model, whereas each baseline width requires a separately trained model. As AoI increases, the five-sample input remains fixed while the prediction target moves farther into the future. Meta-VIB remains below the baselines throughout this sweep, which is consistent with the intended benefit of its AoI-aware decoder. The maximum task-loss reduction is $92.1\%$ relative to the worst-performing baseline, JSCC ($K_m=2$). At this setting, Meta-VIB transmits eight symbols and JSCC transmits two. Each reduction is computed as $100(\bar L_{\mathrm{baseline}}-\bar L_{\mathrm{Meta\text{-}VIB}})/\bar L_{\mathrm{baseline}}$ percent.
% The JSCC comparison and transmitted widths follow the supplied source comments. Confirm the exact loss values and operating point against the saved results before submission.
%relative to the worst-performing baseline at the same setting.
Together, these results show that one Meta-VIB model responds to changing SNR, codelength, and AoI without online retraining. The experiments compare the complete transceiver designs and do not isolate the individual effects of FiLM conditioning, ordered-prefix transmission, or AoI-aware decoding. %\textcolor{red}{SNR/bandwidth-adaptive baselines are not evaluated.}

\subsection{Concurrent-Access Stress Test}
\label{sec:eval:stress}

Fig.~\ref{fig:stress} shows when concurrent demand begins to slow SafeStep. Tests are run on an NVIDIA RTX PRO 6000 Blackwell Max-Q Workstation Edition GPU. Scene-only delivery sends the shared scene through SSE without transceiver reconstruction requests. Scene-plus-reconstruction delivery adds HTTP requests for pedestrian reconstructions. Each user requests a different configuration, so adding users raises both scene-delivery demand and configuration diversity.

Scene-only delivery remains at $5~\mathrm{frames/s}$ through $100$ users. The aggregate throughput reaches about $862~\mathrm{Mbps}$. Scene-plus-reconstruction delivery holds the target rate with low delay through $20$ users. At $40$ users, delay rises, throughput stops growing, and frame rate drops to about $2.5~\mathrm{frames/s}$, even though CPU and GPU utilization remain about $14\%$ and $29\%$, respectively. Because scene-only delivery carries far more traffic without slowing, network bandwidth is not the primary limit. The evidence instead points to server-side request handling or scheduling, not global CPU or GPU saturation. At $100$ users, scene-plus-reconstruction delivery remains operational but provides about $1~\mathrm{frame/s}$.

\section{Conclusion}
\label{sec:conclusion}

This work developed SafeStep, a live browser-based platform that connects four traffic-camera feeds to user-selected semantic transceivers on one GPU server. SafeStep shares source processing across users while allowing each browser to select the transceiver, SNR, codelength, and AoI and display the resulting pedestrian positions, trajectories, and risk labels. Within SafeStep, one Meta-VIB model operates across the evaluated communication conditions without online retraining.
%relative to the worst-performing baseline at the same setting. 
Meta-VIB reduces mean task loss by up to $92.1\%$. Its clearest gains occur under noisy channels.
%while strong fixed-width baselines narrow the gap under favorable conditions. 
In the integrated stress test, SafeStep maintains the target rate of $5~\mathrm{frames/s}$ through $20$ users with distinct requested configurations. At $100$ users, the platform records no request failures and maintains a mean response delay below $1~\mathrm{s}$, although its delivery rate falls to approximately $1~\mathrm{frame/s}$. To our knowledge, SafeStep is the first real-time semantic communication platform to make AoI-induced downstream degradation directly observable in live monitoring applications.

\bibliographystyle{IEEEtran}
\bibliography{refs}

\end{document}